\documentclass[lettersize,journal]{IEEEtran}
\usepackage{amsmath,amsfonts}
\usepackage{algorithmic}
\usepackage{algorithm}
\usepackage{array}
\usepackage{cite}
\usepackage{textcomp}
\usepackage{stfloats}
\usepackage{url}
\usepackage{verbatim}
\usepackage{graphicx}
\usepackage{cite}
\usepackage{tabularx}
\usepackage{booktabs}
\usepackage{bm}
\usepackage{subfigure}
\usepackage{orcidlink} %调包
\usepackage{makecell}
\begin{document}

\title{HERO: Rethinking Visual Token Early Dropping in High-Resolution Large Vision-Language Models}

\author{Xu Li\textsuperscript{$\ast$}$^{\orcidlink{0009-0001-2431-3410}}$,Yuxuan Liang\textsuperscript{$\ast$}$^{\orcidlink{0009-0004-5039-7568}}$, Xiaolei Chen$^{\orcidlink{0009-0005-5700-9326}}$,  Yi Zheng$^{\orcidlink{0009-0006-1549-6979}}$,Haotian Chen$^{\orcidlink{0009-0001-0593-5281}}$, \\ Zhe Liu$^{\orcidlink{0009-0003-3668-6518}}$, Rui Zhu$^{\orcidlink{0009-0007-7226-6923}}$, Bin Li$^{\orcidlink{0000-0002-9633-0033}}$, Xiangyang Xue\textsuperscript{$\dagger$}$^{\orcidlink{0000-0002-4897-9209}}$, Member, IEEE
        % <-this % stops a space
%\thanks{Manuscript received xx xx, xx; revised xx xx, xx. This work is supported by Shanghai Key Laboratory of Intelligent Information Processing and School of Computer Science, Fudan University. (Corresponding author: Xiangyang Xue.)}

% \thanks{Y Liang, X Li, X. Chen, H. Chen, Y Zheng, C Lai, B Li and X. Xue are with the School of Computer Science, Fudan University, Shanghai 200433, China (E-mails: yxliang23@m.fudan.edu.cn, xu\_li23@m.fudan.edu.cn, chenxl23@m.fudan.edu.cn, htchen24@m.fudan.edu.cn, yizhengplus@gmail.com, chlai21@m.fudan.edu.cn, libin@fudan.edu.cn, xyxue@fudan.edu.cn)}

\thanks{X. Li, Y. Liang, X. Chen, H. Chen, Y. Zheng, Z. Liu, R. Zhu, B. Li and X. Xue are with the College of Computer Science and Artificial Intelligence, Fudan University, Shanghai 200433, China (E-mails: {xu\_li23, yxliang23, chenxl23, htchen24, zhengy23, zheliu24, rzhu24, libin, xyxue}@m.fudan.edu.cn). \textbf{$\ast$ denotes equal contribution. $\dagger$denotes the corresponding author}.}

}

\markboth{Journal of \LaTeX\ Class Files,~Vol.~14, No.~8, August~2021}%
{Shell \MakeLowercase{\textit{et al.}}: A Sample Article Using IEEEtran.cls for IEEE Journals}


\maketitle

\begin{abstract}
By cropping high-resolution images into local tiles and encoding them independently, High-Resolution Large Vision-Language Models (HR-LVLMs) have demonstrated remarkable fine-grained visual understanding capabilities. However, this divide-and-conquer paradigm substantially increases the number of visual tokens, leading to significant computational and memory overhead. To better understand and mitigate this challenge, we empirically study visual token utilization in HR-LVLMs and uncover three key findings: (1) the importance of local tiles is jointly determined by visual saliency and textual relevance; (2) the CLS token in CLIP-based vision encoders follows a two-stage attention pattern across layers, with each stage attending to different types of visual tokens; and (3) tokens emphasized at different stages encode information at varying levels of granularity, playing complementary roles within LVLMs. Guided by these insights, we introduce HERO, a High-resolution visual tokEn eaRly drOpping framework. HERO estimates tile-level importance and selectively retains complementary tokens, achieving superior efficiency–accuracy trade-offs across diverse benchmarks and model scales, all in a training-free manner. This work provides both empirical insights and a practical solution for improving inference efficiency in HR-LVLMs.
\end{abstract}


\section{Introduction}

\IEEEPARstart{L}{arge} Vision-Language Models (LVLMs) are a class of multimodal Large Language Models (LLMs) that can tackle both vision-only and vision-language tasks through instruction following \cite{tmm1, tmm2, tmm3}. They typically comprise three core components: a vision encoder, a vision-language adapter, and an LLM. The vision encoder extracts visual features from input images, the adapter projects these features into the LLM’s embedding space, and the LLM integrates them with user instructions to generate task-specific responses.

Most LVLMs adopt CLIP-pretrained Vision Transformers (CLIP-ViTs) \cite{clip} as the vision encoder. These encoders offer two advantages: their serialized outputs naturally conforms to the LLM’s input format, and their image-text contrastive pretraining facilitates alignment with the LLM. However, CLIP-ViTs usually operate at fixed and limited input resolutions (e.g., $336^2$). Consequently, early LVLMs \cite{llava1.5, blip2} have to downscale high-resolution images to fit the constraint, leading to the loss of fine-grained visual details. 

To overcome the limitation, researchers have developed High-Resolution LVLMs (HR-LVLMs), which incorporate a tile cropping strategy during image preprocessing \cite{llavanext, chen2024internvl}. Specifically, the input image is resized and partitioned into local tiles, each matching the encoder's input resolution. In parallel, a downsized version of the original image is generated as a global thumbnail to retain holistic context. These images are encoded independently, and their features are concatenated in sequence while entering the LLM. While this design provides high-resolution visual tokens for the LLM, it significantly increases the token count. Given the quadratic complexity of LLMs with input token length, HR-LVLMs become computationally expensive and difficult to deploy in resource-constrained scenarios.

Visual token compression has become a popular topic in LVLM research. Early approaches address this challenge by designing vision-language adapters with feature abstraction capabilities \cite{blip2,c-abstractor}. However, these methods require model training, thus incurring high deployment costs and poor generalizability. Subsequent works shift toward inference-time token pruning, where initial methods remove visual tokens within the LLM based on text-to-image attention \cite{fastv, sparsevlm}. While these approaches offer plug-and-play flexibility, they may interfere with efficient LLM inference mechanisms such as KV caching \cite{KVcache} and Flash-Attention \cite{flashattention}. Therefore, more recent studies propose the concept of early dropping \cite{llava-prumerge, fastervlm}, which removes low-contributing visual tokens before they enter the LLM, typically based on CLS attention in the vision encoder. However, most of the early dropping methods are designed for conventional LVLMs and do not account for the specific structure of HR-LVLMs.

Until recently, GlobalCo$\mathrm{m}^2$ \cite{globalcom2} and HiRED \cite{hired} extend the philosophy of visual token early dropping to HR-LVLMs via a two-step pipeline. First, they allocate a constrained token budget across local tiles by estimating regional importance from the CLS attention of the global thumbnail. Next, they select a subset of visual tokens to retain within each tile according to the tile’s own CLS attention. While effective, these methods have three limitations: (1) token budget allocation overlooks the semantic relevance between each tile and the instruction, and is further affected by spatial misalignment introduced by resizing and padding during tile cropping; (2) the choice of attention layer relies on sparse visualizations or limited ablations, lacking a layerwise analysis of attention dynamics in CLIP-ViTs; (3) token selection is applied uniformly across local tiles and the global thumbnail, ignoring their complementary roles in preserving fine-grained details and global context.

To address these limitations, we conduct a series of pilot experiments and uncover three key insights:
(1) the importance of local tiles are jointly decided by their visual saliency and textual relevance;
(2) the CLS attention in CLIP-ViTs exhibits a two-stage pattern across layers, where each stage focuses on different image patches; and
(3) visual tokens emphasized at the two stages encode information at varying levels of granularity, playing complementary roles within LVLMs.
Motivated by these insights, we propose HERO, a unified visual token early dropping framework for HR-LVLMs. By estimating tile importance from a multimodal perspective and retaining complementary visual tokens across local and global views, HERO enables HR-LVLMs to achieve faster inference with minimal performance loss or even with performance gains. Our main contributions are concluded as follows:
\begin{itemize}
\item We conduct systematic empirical analysis on three underexplored aspects of HR-LVLMs: (i) reliable estimation of tile importance, (ii) layerwise dynamics of CLS attention in CLIP-based vision encoders, and (iii) complementary functions of visual tokens within LLMs.
\item We propose HERO, a visual token early dropping method that supports multimodal-guided token budget allocation and function-aware token selection within HR-LVLMs.
\item We integrate HERO into representative HR-LVLMs of different model scales and comprehensively evaluate it on diverse benchmarks, demonstrating consistent improvements over existing state-of-the-art methods. 
\end{itemize} 

\section{Related Work}

\subsection{Development of LVLMs}
By integrating pretrained vision encoders with LLMs, LVLMs can jointly interpret visual and textual inputs. Most LVLMs employ a CLIP-ViT to obtain semantically rich visual features. However, these vision encoders usually operate at fixed and limited input resolutions. This requires early LVLMs \cite{blip2, llava1.5} to downscale high-resolution images before encoding, which impairs their ability to capture fine-grained visual details. To address this limitation, HR-LVLMs \cite{llavanext,chen2024internvl} are proposed. These models resize and partition the input image into local tiles, each matching the encoder's input resolution, and simultaneously generate a downsized global thumbnail to preserve global context. All tiles and the thumbnail are encoded independently, and their features are concatenated in the sequence dimension before being passed to the LLM. While this strategy effectively preserves detailed visual information, it substantially increases the number of visual tokens, which significantly increases computational complexity and memory consumption. These limitations of HR-LVLMs naturally motivate the exploration of efficient token compression mechanisms tailored for high-resolution inputs.

\subsection{Visual Token Compression in LVLMs}
To mitigate computational overhead, visual token compression strategies have been developed for LVLMs. Early approaches focus on designing novel adapters, such as Q-Former \cite{blip2} and C-Abstractor \cite{c-abstractor}, to distill visual features into a compact token set. However, these methods require model training, which increases deployment costs and hinders generalizability. To address these issues, inference-time token pruning methods are proposed. FastV \cite{fastv} and SparseVLM \cite{sparsevlm} rank visual tokens based on text-to-image attention within the LLM and discard those with low relevance. They avoid model training but can interfere with LLM inference mechanisms like KV caching \cite{KVcache} and Flash-Attention \cite{flashattention}. Therefore, early dropping strategies are proposed to prune visual tokens before the LLM. PruMerge \cite{llava-prumerge} and FasterVLM \cite{fastervlm}  discard uninformative visual tokens based on CLS attention from the vision encoder. However, they are designed for conventional LVLMs without considering the characteristics of HR-LVLMs. To address this, GlobalCo$\mathrm{m}^2$ \cite{globalcom2} and HiRED \cite{hired} are introduced. These methods first allocate token budgets across local tiles using the CLS attention from the global thumbnail, then select informative tokens within each tile based on local CLS attention. Although tailored for HR-LVLMs, they still exhibit limitations such as insufficient textual relevance modeling, heuristic attention extraction, and uniform token selection, highlighting the need for further innovation.

\section{Empirical Observations}
\label{priliminary}
Motivated by the limitations of prior methods, we conduct empirical analysis on three aspects of HR-LVLMs: (1) the measurement of tile importance, (2)  the layerwise dynamics of CLS attention in CLIP-ViTs, and (3) the distinct roles of visual tokens highlighted by CLS attention at different layers.

\subsection{Measurement of Tile Importance}
\label{preliminary_tile_importance}
Estimating the relative importance of local tiles is crucial for allocating a limited token budget to the most informative regions. Prior methods typically derive tile importance from the CLS attention of the global thumbnail. However, this strategy has two limitations: (i) it overlooks textual relevance—tiles that are not globally salient yet highly related to the instruction can be under-allocated; and (ii) it suffers from spatial misalignment—the thumbnail's attention map does not precisely align with the tile grids due to resizing and padding. We argue that a reliable estimator should integrate both visual and textual cues and be computed directly at the tile level. 

\begin{figure}[ht]
\centering
\includegraphics[width=0.45\textwidth]{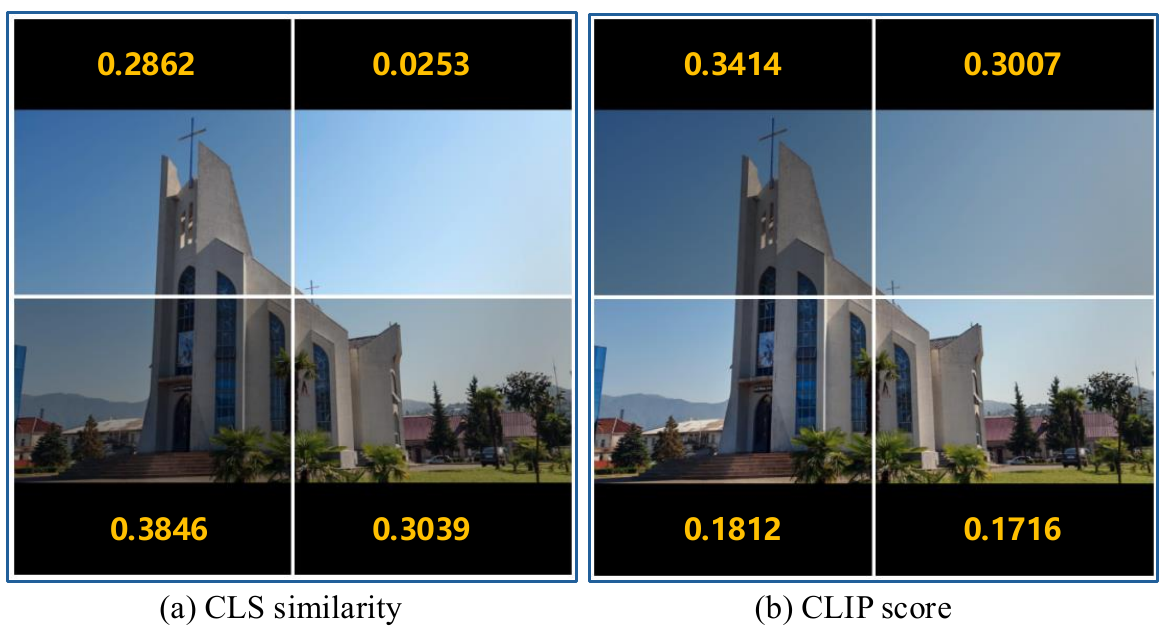}
\caption{Illustration of tile-level importance estimation. For each local tile, we report its CLS similarity to the global thumbnail (reflecting visual saliency) and its CLIP score with the instruction “How many crucifixes are there?” (reflecting textual relevance). All scores are normalized across tiles.}
\label{tile_importance}
\end{figure}

Leveraging the properties of CLIP-ViTs, two tile-level metrics can be naturally defined: (1) the cosine similarity between a tile’s CLS token and that of the global thumbnail (CLS similarity), which reflects its visual saliency from a global perspective; and (2) the cosine similarity between a tile’s CLS token and the textual embedding of the instruction (CLIP score), which measures its textual relevance. As illustrated in Fig.~\ref{tile_importance}, CLS similarity highlights tiles with prominent structures (e.g., the church body), whereas the CLIP score is more sensitive to instruction-aligned regions (e.g., the two crucifixes on the roof), indicating complementary strengths.

To quantitatively assess these metrics, we evaluate LLaVA-NeXT-7B~\cite{llavanext} on four benchmarks of different categories: MME~\cite{mme}, POPE~\cite{pope}, TextVQA~\cite{textvqa}, and RWQA~\cite{realworldqa}. We first measure performance using only the global thumbnail. Then, we add exactly one local tile selected by ranking tiles individually with either CLS similarity (to the thumbnail) or CLIP score (to the instruction), and record the performance variation. As shown in Table~\ref{combined}, adding the top-ranked tile consistently yields the largest gains, whereas lower-ranked tiles can introduce irrelevant or noisy content and even degrade accuracy. These results validate both metrics and motivate combining them for more reliable tile-importance estimation.

\begin{table}[t]
\centering
\small
\caption{Performance of combining the global thumbnail with a single local tile. “+Top-$k$” denotes the global thumbnail plus the $k$-th ranked tile by CLS similarity or CLIP score, with no cumulative addition of tiles.}
\label{combined}
\setlength{\tabcolsep}{3pt}
\renewcommand{\arraystretch}{0.9}  % 调整行间距
\begin{tabular}{lcccc}
\toprule
\textbf{Setting} & \textbf{MME} & \textbf{POPE} & \textbf{TextVQA} & \textbf{RWQA} \\
\midrule
Global Thumbnail                      
    & 1490.9 
    & 84.7 
    & 54.1 
    & 51.7 \\
\midrule
+Top-1 CLIP score        
    & \textbf{1523.1} 
    & \textbf{87.1} 
    & \textbf{57.1} 
    & \textbf{55.0} \\
+Top-2 CLIP score      
    & \underline{1500.5} 
    & \underline{86.2} 
    & \underline{54.6} 
    & \underline{52.9} \\
+Top-3 CLIP score       
    & 1489.1 
    & 85.7 
    & 53.6 
    & 52.8 \\
+Top-4 CLIP score      
    & 1477.9 
    & 85.2 
    & 53.0 
    & 52.6 \\
\midrule
+Top-1 CLS similarity   
    & \textbf{1529.2} 
    & \textbf{86.8} 
    & \textbf{55.7} 
    & \textbf{53.4} \\
+Top-2 CLS similarity
    & \underline{1512.6} 
    & \underline{86.5} 
    & \underline{54.5} 
    & \underline{53.3} \\
+Top-3 CLS similarity 
    & 1479.0 
    & 86.1 
    & 54.1 
    & 53.1 \\
+Top-4 CLS similarity
    & 1467.7 
    & 85.7 
    & 54.0 
    & 52.4 \\
\bottomrule
\end{tabular}

\end{table}

\subsection{Dynamics of CLS Attention in CLIP-ViTs}
\label{subsection_attn_dynamics}
In CLIP-ViTs, the CLS token is used to aggregate global information, and its outgoing attention weights naturally indicate the relative importance of patch tokens. Prior methods \cite{fastervlm, globalcom2, hired} have exploited this property to guide token selection for both local tiles and the global thumbnail. However, they typically focus on either the earliest or the final layers, while overlooking intermediate layers that often capture richer and more diverse semantic information \cite{igva}.

\begin{figure}[t]
\centering
\includegraphics[width=0.5\textwidth]{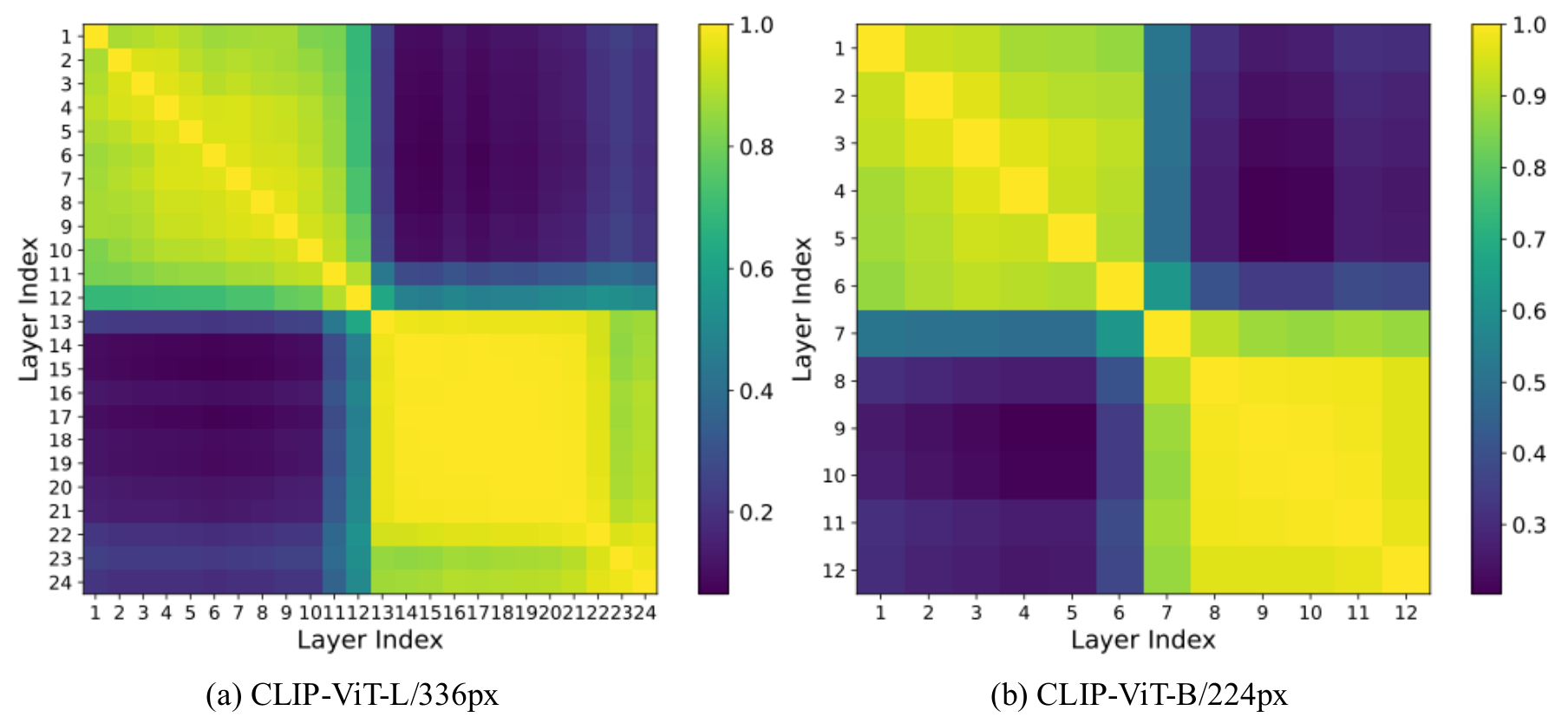}
\caption{Cosine similarity between CLS attention maps across different layers of CLIP-ViTs. The similarities are averaged over attention heads and computed using all images from the MME benchmark.}
\label{attention_similarity}
\end{figure}

We first investigate the pairwise cosine similarity of CLS attention maps across different layers for two variants of CLIP-ViTs, using images from the MME benchmark. As shown in Fig.~\ref{attention_similarity}, both models exhibit a clear two-stage pattern. Specifically, attention maps within the early-to-middle layers (approximately layers 1–12 in CLIP-ViT-L and 1–6 in CLIP-ViT-B) are highly similar to each other, while those within the later layers (layers 13–24 in CLIP-ViT-L and 7–12 in CLIP-ViT-B) also form a highly coherent cluster. In contrast, the cross-stage similarity between the two groups is significantly lower, indicating a sharp transition in CLS attention behavior around the midpoint of the network.

Building on the two-stage pattern identified above, we further visualize CLS attention layer by layer in CLIP-ViT-L/336px on two representative images: one with salient foreground objects and another with less prominent foreground and a cluttered background. As shown in Fig.~\ref{ship} and Fig.~\ref{street} of Appendix~\ref{appendix_layerwise}, several consistent trends can be observed: (1) in the initial layers, CLS attention is broadly distributed and often attracted by low-level textures and background structures such as sky, water surfaces, or road edges, resulting in noisy patterns; (2) from the early to middle layers, CLS attention consolidates onto the primary visual objects, but as the network approaches the midpoint, the attention begins to diffuse; (3) beyond this midpoint, a sharp transition occurs: CLS attention collapses onto a small subset of background tokens, and this highly concentrated focus persists until the final layers, where it is slightly relaxed.

These observations provide two important insights. First, to capture the main semantic content, CLS attention from early-to-middle layers is generally more reliable than that from the initial or final layers. Second, relying on a single layer is insufficient, since the optimal layer varies across image samples. For instance, in the ship example (Fig.~\ref{ship}), layer 4 already provides good coverage of the main objects, whereas in the street scene (Fig.~\ref{street}), layer 9 offers the most complete focus on the relevant foreground content.

\subsection{Roles of Different Visual Tokens in LVLMs}
\label{subsection_token_roles}

\begin{table}[t]
\centering
\small
\caption{Performance comparison on MMStar. C.P. denotes coarse perception, F.P. denotes fine-grained perception, I.R. denotes Instance Reasoning, L.R. denotes Logical Reasoning, M.A. denotes math, and S.T. denotes science and technology.}
\label{token_type_comparison_mmstar}
\setlength{\tabcolsep}{3pt}
\renewcommand{\arraystretch}{0.9}  % 调整行间距
\begin{tabular}{lcccccc}
\toprule
\textbf{Token Selection} & \textbf{C.P.} & \textbf{F.P.} & \textbf{I.R.} & \textbf{L.R.} & \textbf{M.A.}  & \textbf{S.T.}\\
\midrule
Setting-1                      
    & 57.0 
    & 24.4 
    & 39.5 
    & 27.0
    & 29.4
    & 20.2\\
% \midrule
Setting-2        
    & 58.0
    & 22.4
    & 38.6
    & 28.5
    & 28.5
    & 23.8\\
\bottomrule
\end{tabular}

\end{table}

\begin{table}[t]
\centering
\small
\caption{Image captioning scores (1-5 scale) provided by GPT-4o on different evaluation dimensions.}
\label{token_type_comparison}
\setlength{\tabcolsep}{4pt}
\renewcommand{\arraystretch}{0.8}
\begin{tabular}{lccc}
\toprule
\textbf{Token Selection} & \textbf{Object Coverage} & \textbf{Object Correctness} & \textbf{Avg.} \\
\midrule
Setting-1   & 3.8 & 3.6 & 3.7 \\
Setting-2   & 3.4 & 1.9 & 2.7 \\
\midrule
\textbf{Token Selection} & \textbf{Global Coverage} & \textbf{Global Correctness} & \textbf{Avg.} \\
\midrule
Setting-1   & 2.9 & 1.9 & 2.4 \\
Setting-2   & 3.4 & 2.3 & 2.9 \\
\bottomrule
\end{tabular}

\end{table}
Building on the observations above, we define two types of visual tokens in CLIP-ViTs: (1) primary tokens, which are spatially aligned with the main semantic content of the image and receive most CLS attention in the early-to-middle layers; and (2) shortcut tokens, which are typically located in background regions but attract the majority of CLS attention in the middle-to-late layers.

To examine whether these two types of tokens play distinct roles in LVLMs, we design two inference settings for LLaVA-v1.5-7B~\cite{llava1.5}: (1) retaining only the top 10\% of CLS-most-attended tokens from layers 4–8 (i.e., primary tokens), and (2) retaining only the top 10\% from layers 12–23 (i.e., shortcut tokens). We first evaluate both settings on the MMStar benchmark~\cite{mmstar}. As shown in Table~\ref{token_type_comparison_mmstar}, setting-1 surpasses setting-2 in categories requiring precise grounding of local details, such as fine-grained perception, instance reasoning, and math. In contrast, setting-2 performs better in categories that benefit from holistic cues and higher-level reasoning, including coarse perception and logical reasoning. We further assess the two settings through descriptive generation. For each image in MMStar, we generate a caption under each setting and ask GPT-4o to score the captions along four dimensions (see Appendix ~\ref{appendix_caption} for scoring definitions and prompts). As shown in Table~\ref{token_type_comparison}, setting-1 achieves higher object-level scores, while setting-2 achieves higher global-level scores. These consistent trends suggest that primary tokens are more effective for fine-grained, object-centric understanding, whereas shortcut tokens contribute more to tasks that rely on global or abstract context.

Overall, these findings suggest a functional division: primary tokens are best retained in local tiles to preserve fine-grained semantics, while shortcut tokens are better retained in the global thumbnail to convey holistic context.

\section{Method}

\begin{figure*}[h]
\centering
\includegraphics[width=1\textwidth]{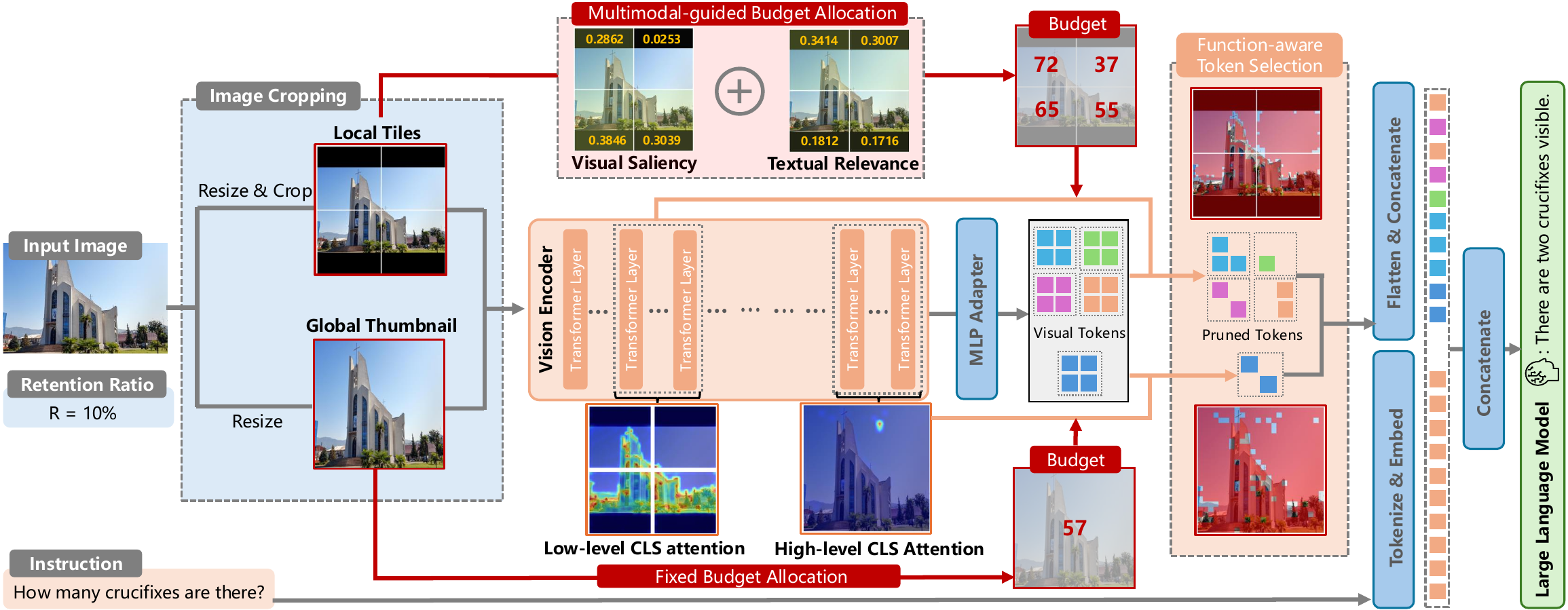}
\caption{Overview of HERO's workflow. Red arrows indicate the process of token budget allocation, orange arrows denote the process of token selection, and gray arrows illustrate the original inference pipeline of the baseline HR-LVLM.}
\label{model}
\end{figure*}

Motivated by the empirical observations in Section~\ref{priliminary}, we propose HERO, a visual token early dropping framework tailored for HR-LVLMs. In this section, we first review the baseline model, then present the integration of HERO, and finally analyze the theoretical efficiency.

\subsection{Baseline Model}
We adopt the LLaVA-NeXT series~\cite{llavanext} as the baseline HR-LVLMs due to both its representativeness and widespread use in recent studies. 

\subsubsection{Model Architecture}
LLaVA-NeXT model family follows the standard LVLM architecture, employing CLIP-ViT-L/336px as the vision encoder, a two-layer MLP as the vision-language adapter, and Vicuna-v1.5 \cite{vicuna} as the LLM.

\subsubsection{Inference Pipeline} To preserve fine-grained details in high-resolution images, LLaVA-NeXT adopts an image tiling strategy. Specifically, it defines five target resolutions: $\{$(336, 672), (672, 336), (672, 672), (1008, 336), (336, 1008)$\}$, corresponding to five tiling grids: $\{$(1, 2), (2, 1), (2, 2), (3, 1), (1, 3)$\}$. For each input image, the resolution with the closest aspect ratio is selected. The image is first padded along the shorter side to match the aspect ratio, and then resized to the target resolution. The resized image is divided into non-overlapping tiles, each of size $336^2$, matching the input resolution of the vision encoder. Meanwhile, the original image is also resized to $336^2$ to produce a global thumbnail that preserves holistic context. The local tiles and the global thumbnail are independently encoded by the vision encoder, each yielding 576 patch embeddings plus one CLS embedding. The patch embeddings are projected through the MLP adapter to form visual tokens. Finally, the visual tokens are concatenated in sequence with the instruction embeddings and fed into the LLM for multimodal reasoning and response generation.

\subsection{Design of HERO}

As illustrated in Fig.~\ref{model}, HERO operates in two stages. First, it adaptively allocates a user–specified visual token budget across local tiles and the global thumbnail. Second, it retains the top tokens within each region (tile or thumbnail) according to token informativeness and the distinct roles of local versus global views. The framework is entirely training–free and can be used as a drop–in component for HR-LVLMs.

\subsubsection{Token Budgeting}
To accommodate different computational constraints, HERO allows users to specify a target visual token retention ratio $R \in (0, 1]$. Suppose an input image is partitioned into $K$ local tiles plus one global thumbnail, each yielding $N$ visual tokens. The total budget is $N' = \lfloor(K+1)\cdot N\cdot R\rfloor$. We first reserve a fixed quota of $N_G = \lfloor N\cdot R \rfloor$ tokens for the thumbnail to guarantee global context, and distribute the remaining $N_L = N' - N_G$ tokens among the local tiles in a content-adaptive manner.

Motivated by the analysis in Section~\ref{preliminary_tile_importance}, HERO scores each tile by jointly considering visual saliency and textual relevance. Let $\mathrm{CLSsim}_i$ denote the cosine similarity between the CLS embedding of tile $i$ and that of the global thumbnail. 
The visual saliency score of tile $i$ is defined as:
\begin{equation}
    s^v_i = \frac{\mathrm{exp}(\mathrm{CLSsim}_i)}{\sum_{j=1}^K\mathrm{exp}(\mathrm{CLSsim}_j)} \in [0, 1], \quad i \in \{1,2,..K\}.
\label{eq-tv}
\end{equation}
This score reflects the relative saliency of each tile's visual content from a global perspective. 

For textual relevance, HERO leverages the CLIP image–text similarity between each tile and the instruction. Let $\mathrm{CLIPscore}_i$ denote this similarity for tile $i$.  The textual relevance score is defined as:
\begin{equation}
s^t_i = \frac{\mathrm{exp}(\mathrm{CLIPscore}_i)}{\sum_{j=1}^K\mathrm{exp}(\mathrm{CLIPscore}_j)} \in [0, 1], \quad i \in \{1,2,..K\}.
\label{eq-tt}
\end{equation}
This score captures the relative alignment between the visual content of each tile and the user-provided instruction. 

The overall importance of a tile is then computed as a convex combination of the two scores: 
\begin{equation}
s_i = \alpha \cdot s^v_i + (1-\alpha) \cdot s^t_i, \quad i \in \{1,2,..K\},
\label{eq-alpha}
\end{equation}
where $\alpha \in [0, 1]$ is a hyperparameter that controls the balance between visual saliency and textual relevance. Finally, the number of tokens allocated to each tile is determined by:
\begin{equation}
N_{L_i}=\lfloor N_L\cdot s_i\rfloor, \quad i \in \{1,2,..K\}.
\end{equation}

This budgeting process is performed entirely at the tile level: it fuses vision-guided and text-guided signals, requires no patch-level attention extraction, and avoids cross-scale grid alignment issues present in prior methods.

\subsubsection{Token Selection}
After allocating token budgets, HERO selects the most informative tokens to retain within each local tile and the global thumbnail. 

As discussed in Sections~\ref{subsection_attn_dynamics} and~\ref{subsection_token_roles}, CLS attention in CLIP-ViTs emphasizes different types of visual tokens at different depths: in low-to-middle layers, the CLS token predominantly attends to primary tokens that capture fine-grained semantics, whereas in higher layers, it shifts toward shortcut tokens that encode global context. To exploit this property, HERO defines two sets of layer indices: $\mathcal{L}$ for low-to-middle layers and $\mathcal{H}$ for middle-to-high layers, and leverages CLS attention from these sets to guide token selection in a function-aware manner.

For the global thumbnail, which is intended to convey scene-level semantics, HERO estimates token importance by averaging CLS attention across the layers in $\mathcal{H}$: 
\begin{equation}
a^G_j = \frac{1}{|\mathcal{H}|} \sum _{l \in \mathcal{H}} a_j^{G, l}, \quad j \in \{1, 2, .., N\},
\label{eq-H}
\end{equation}
where $a_j^{G, l}$ denotes the attention weight from the CLS token to patch token $j$ in layer $l$, averaged across all attention heads. The top-$N_G$ tokens within the thumbnail are retained according to this score.

For each local tile, whose goal is to preserve fine-grained details, HERO computes token importance by averaging CLS attention across the layers in $\mathcal{L}$:
\begin{equation}
a_j^{L_i} = \frac{1}{|\mathcal{L}|} \sum _{l \in \mathcal{L}} a_j^{L_i, l}, \quad j \in \{1, 2, ..N\},
\label{eq-L}
\end{equation}
where $a_j^{L_i, l}$ denotes the CLS attention weight from the CLS token to patch token $j$ in tile $i$ at layer $l$, averaged across all heads. The top-$N_{L_i}$ tokens within each tile are then retained.

This selection strategy offers two key advantages. First, it applies distinct criteria to the global thumbnail and local tiles, aligning with their complementary roles in HR-LVLMs. Second, by aggregating CLS attention across multiple layers in $\mathcal{L}$, it enhances robustness and adaptability, ensuring that the primary semantic content is consistently captured across diverse visual inputs.

\subsection{Theoretical Efficiency Analysis}
In LVLMs, the LLM constitutes the primary source of computational overhead. 
During inference, the LLM processes all input tokens in a single forward pass (the prefill stage) to cache intermediate states for subsequent decoding. Since every token participates in self-attention during this stage, the prefill computation dominates the overall complexity. The per-layer FLOPs of the LLM in the prefill stage can be expressed as:
\begin{equation}
8(N_v+N_t)d^2 + 4(N_v+N_t)^2d + 6(N_v+N_t)dm,
\end{equation}
where $N_v$ and $N_t$ denote the numbers of visual and textual tokens, $d$ is the hidden dimension of the LLM, and $m$ is the intermediate dimension of its MLP layers. 
Because $N_v$ usually far exceeds $N_t$, visual tokens represent the dominant computational bottleneck. 

By applying HERO with a token retention ratio $R \in (0,1]$, the number of visual tokens is reduced from $N_v$ to $RN_v$, and the per-layer FLOPs become:
\begin{equation}
8(RN_v+N_t)d^2 + 4(RN_v+N_t)^2d + 6(RN_v+N_t)dm.
\end{equation}
In typical scenarios where $N_v \gg N_t$, this reduction substantially decreases the quadratic term, leading to marked improvements in computational efficiency. 
Hence, HERO achieves efficient inference in HR-LVLMs by directly lowering the dominant cost of processing visual tokens.

\section{Experiments}

\subsection{Experimental Setting}
\subsubsection{Implementation Details}
We integrate HERO into both LLaVA-NeXT-7B and LLaVA-NeXT-13B. For token budget allocation, we set $\alpha=0.5$. For token selection, we set $\mathcal{L} = \{6, 7, 8, 9, 10\}$ and $\mathcal{H} = \{22\}$. All experiments are conducted on a cluster of eight NVIDIA L20 GPUs.

\subsubsection{Benchmarks}
We evaluate HERO on ten benchmarks spanning a wide range of tasks: VQAv2 \cite{vqa-v2}, VizWiz \cite{vizwiz}, and GQA \cite{gqa} for general visual understanding; SQA \cite{sqa} and ChartQA \cite{chartqa} for multimodal reasoning; TextVQA \cite{textvqa} and OCRBench \cite{ocrbench} for text recognition; MME \cite{mme} and MMB \cite{mmb} for comprehensive evaluation; and POPE \cite{pope} for hallucination assessment.

\subsubsection{Comparison Methods}
We compare HERO with three mainstream categories of visual token compression methods: (1) methods that drop visual tokens within the LLM: FastV \cite{fastv} and SparseVLM \cite{sparsevlm}; (2) methods that apply early dropping but are not tailored for HR-LVLMs: LLaVA-Prumerge \cite{llava-prumerge} and FasterVLM \cite{fastervlm}; (3) methods that apply early dropping and explicitly designed for HR-LVLMs: HiRED \cite{hired} and GlobalCo$\mathrm{m}^2$ \cite{globalcom2}. 

\subsection{Main Results}

\begin{table*}[ht]
\centering
\small
\caption{Performance comparison on LLaVA-NeXT-7B and LLaVA-NeXT-13B under different token budget settings.}
\label{performance}
\setlength{\tabcolsep}{5pt} % 缩小列间距
\renewcommand{\arraystretch}{0.8}  % 调整行间距
\begin{tabular}{l|cccccccccc}
\toprule
\textbf{Method (Token Budget)} & \textbf{VQA-v2} & \textbf{VizWiz} & \textbf{GQA} & \textbf{SQA} & \textbf{ChartQA} & \textbf{TextVQA} & \textbf{OCRBench} & \textbf{MME} & \textbf{MMB} & \textbf{POPE} \\
\midrule
LLaVA-NeXT-7B (100\%)  & 81.8 & 57.6 & 64.2 & 70.1 & 54.8 & 64.8 & 50.1 & 1519.0 & 67.4 & 87.6 \\
\midrule % 在这里加一条直线
+ Prumerge+ (55\%)   & 78.0 & --   & --   & 68.2 & 30.2 & 54.4 & 36.5 & 1474.0 & --   & 87.9 \\
+ FastV (50\%)           & 80.7 & 54.9 & 61.8 & 69.1 & --   & 59.6 & --  & 1490.3 & 67.4 & 85.5 \\
+ SparseVLM (50\%)       & \underline{80.9} & 55.7 & 62.0 & 68.1 & --   & 60.0 & --  & 1484.9 & 65.7 & 73.4 \\
+ FasterVLM (50\%)       & 80.6 & 56.4 & 63.4 & 69.1 & --   & 58.9 & --  & 1533.3 & 67.4 & 87.7 \\
+ GlobalCo$\mathrm{m}^2$ (50\%)      & 80.6 & \underline{56.5} & \underline{63.9} & 68.5 & --   & 59.5 & --  & \textbf{1552.9} & \underline{67.6} & 88.1 \\
+ HiRED (40\%)           & 78.8 & --   & --   & \underline{70.8} & \underline{46.5} & \underline{63.6} & \underline{48.8} & 1474.0 & --   & \underline{88.2} \\
+ \textbf{HERO (40\%)}            & \textbf{81.3} & \textbf{57.3} & \textbf{64.3} & \textbf{72.2} & \textbf{48.5} & \textbf{64.5} & \textbf{52.4} & \underline{1550.8} & \textbf{67.9} & \textbf{88.9} \\
\midrule
+ FastV (25\%)           & 78.9 & 54.2 & 60.4 & \underline{69.8} & --   & 58.4 & --  & 1477.3 & 65.6 & 83.1 \\
+ SparseVLM (25\%)       & 78.9 & 55.6 & 60.9 & 67.5 & --   & 58.1 & --  & 1446.1 & 63.8 & 71.0 \\
+ FasterVLM (25\%)       & 78.3 & 55.4 & 61.3 & 67.1 & --   & 58.8 & --  & 1454.6 & \underline{66.0} & 87.2 \\
+ GlobalCo$\mathrm{m}^2$ (25\%)      & \underline{79.4} & \underline{55.7} & \underline{61.4} & 68.1 & --   & 59.2 & --  & \underline{1493.5} & 65.9 & \underline{87.6} \\
+ HiRED (20\%)           & 77.5 & --   & --   & \underline{70.4} & \underline{42.0} & \underline{61.4} & \underline{47.5} & 1483.0 & --   & 87.0 \\
+ \textbf{HERO (20\%)}            & \textbf{79.8} & \textbf{56.4} & \textbf{62.7} & \textbf{71.9} & \textbf{42.8} & \textbf{61.9} & \textbf{48.6} & \textbf{1500.9} & \textbf{66.9} & \textbf{88.4} \\
\midrule
\midrule
% \toprule
% \bottomrule

LLaVA-NeXT-13B (100\%)  & 82.8 & 60.5 & 65.4 & 73.5 & 66.2 & 66.9 & 50.8 & 1572.0 & 70.0 & 87.1 \\
\midrule
+ Prumerge+ (55\%)   & 79.1 & --   & --   & 70.7 & 31.0 & 55.9 & 38.1 & 1480.0 & --   & 87.5 \\
+ FasterVLM (50\%)       & \underline{81.3} & 57.0 & 64.2 & 72.5 & --   & 62.4 & --  & 1534.1 & \underline{69.5} & 87.6 \\
+ GlobalCo$\mathrm{m}^2$ (50\%)      & 81.0 & \underline{57.1} & \underline{64.7} & \underline{73.2} & --   & 62.5 & --  & 1553.5 & 69.3 & \underline{87.7} \\
+ HiRED (40\%)           & 79.3 & --   & --   & \underline{73.2} & \underline{53.7} & \underline{65.2} & \underline{49.1} & \textbf{1570.0} & --   & \underline{87.7} \\
+ \textbf{HERO (40\%)}            & \textbf{81.6} & \textbf{58.0} & \textbf{65.6} & \textbf{73.7} & \textbf{54.4} & \textbf{66.4} & \textbf{52.9} & \underline{1558.6} & \textbf{70.2} & \textbf{88.3} \\
\midrule
+ FasterVLM (25\%)       & 78.9 & 55.0 & 62.3 & 72.1 & --   & 61.2 & --  & 1516.1 & 67.6 & 86.1 \\
+ GlobalCo$\mathrm{m}^2$ (25\%)      & \underline{79.9} & \underline{55.1} & \underline{62.7} & \underline{72.3} & --   & 61.5 & --  & 1531.2 & \underline{67.9} & 86.5 \\
+ HiRED (20\%)           & 77.9 & --   & --   & 71.9 & \underline{48.9} & \underline{63.6} & 46.2 & \underline{1545.0} & --   & \underline{86.7} \\
+ \textbf{HERO (20\%)}            & \textbf{80.6} & \textbf{56.3} & \textbf{63.7} & \textbf{72.8} & \textbf{49.9} & \textbf{64.3} & \textbf{49.4} & \textbf{1548.9} & \textbf{68.0} & \textbf{86.9} \\

\bottomrule
\end{tabular}

\end{table*}

We evaluate HERO under two representative token budget settings: a moderate one ($40\%$) and an aggressive one ($20\%$).

\subsubsection{With LLaVA-NeXT-7B}
The results on LLaVA-NeXT-7B are shown in the upper section of Table~\ref{performance}. 
Under the moderate $40\%$ budget, HERO exhibits strong robustness: it even surpasses the full-token baseline on 6 out of 10 benchmarks, including GQA, SQA, OCRBench, MME, MMB, and POPE, suggesting that pruning uninformative visual tokens can effectively reduce noise in the visual modality. 
Moreover, HERO achieves the best performance among all comparison methods on 9 out of 10 benchmarks, trailing slightly behind GlobalCo$\mathrm{m}^2$ on MME. 
With the more aggressive $20\%$ budget, HERO demonstrates even stronger competitiveness: it not only surpasses the full-token baseline on SQA and POPE, but also secures the best results across all 10 benchmarks. 
These consistent improvements highlight the superiority of HERO in striking a balance between efficiency and accuracy.

\subsubsection{With LLaVA-NeXT-13B}

The results on LLaVA-NeXT-13B are presented in the lower section of Table~\ref{performance}. 
Under the moderate $40\%$ budget, HERO again delivers strong results, surpassing the full-token baseline on GQA, SQA, OCRBench, MMB, and POPE, and achieving the best scores among the comparison methods on 9 out of 10 benchmarks. 
At the aggressive $20\%$ budget, HERO outperforms all competing methods on every benchmark and even surpasses the full-token baseline on POPE. 
Notably, we observe that the performance retention under both token budgets is generally lower for the 13B model compared to the 7B model. 
This may be attributed to the stronger baseline model relying more heavily on comprehensive visual information, making it more sensitive to token removal. 
Nevertheless, HERO consistently ranks as the top-performing method across both model scales, further validating its effectiveness and generality.

\subsection{Ablation Studies}
We conduct ablation studies in two parts: (1) assessing the contribution of each key component of HERO, and (2) exploring the impact of different hyperparameter configurations to identify the optimal settings.

\subsubsection{Component Ablation}
We evaluate the contribution of each component by progressively removing them. As shown in Table \ref{ablation_key}, setting-1 removes textual relevance from the budget allocation, assigning token quotas across tiles solely based on visual saliency; setting-2 further discards visual saliency, distributing tokens uniformly across tiles; setting-3 and setting-4 eliminate the differentiated token selection between tiles and the global thumbnail. Following prior work \cite{globalcom2, hired}, setting-3 relies exclusively on CLS attention from the first layer of CLIP-ViT for both tiles and the thumbnail, whereas setting-4 relies solely on CLS attention from the final layer.

Compared with the final method, setting-1 shows consistent performance drops across all benchmarks, most notably on OCRBench, confirming the necessity of incorporating textual relevance into budget allocation for text-intensive tasks. 
Setting-2 degrades further, particularly on MME and OCRBench, indicating that ignoring both visual saliency and textual relevance prevents the model from prioritizing informative regions. 
Settings-3 and 4 perform worse than setting-2, demonstrating the importance of applying distinct token selection strategies to local tiles and the global thumbnail. 
Notably, when token selection is restricted to a single CLS attention layer, the final layer (setting-4) yields better performance than the first layer (setting-3). 
This suggests that although final-layer CLS attention alone cannot replace HERO’s multi-layer design, it provides more semantically aggregated signals than initial-layer attention, which tends to overemphasize low-level textures and background patterns. 

These results validate the effectiveness of each component in HERO and highlight the importance of both multimodal-guided budget allocation and function-aware token selection.

\begin{table}[h]
\centering
\small
\caption{Ablation results of progressively removing key components from HERO under 20\% token budget with LLaVA-NeXT-7B.}
\label{ablation_key}
\setlength{\tabcolsep}{4pt}
\renewcommand{\arraystretch}{0.9}  % 调整行间距
\begin{tabular}{lcccc}
\toprule
\textbf{Configuration} & \textbf{GQA} & \textbf{SQA} & \textbf{OCRBench} & \textbf{MME} \\
\midrule
HERO      & 62.7 & 71.9 & 52.4 & 1500.9 \\
Setting-1 & 62.3 & 71.7 & 46.1 & 1494.1 \\
Setting-2 & 62.0 & 71.6 & 45.8 & 1488.1 \\
Setting-3 & 60.9 & 71.3 & 42.8 & 1439.8 \\
Setting-4 & 61.8 & 71.5 & 45.4 & 1472.6 \\
\bottomrule
\end{tabular}

\end{table}

\begin{table}[t]
\centering
\small
\caption{Ablation results of $\mathcal L$. C.P. denotes coarse perception, F.P. denotes fine-grained perception, I.R. denotes Instance Reasoning, L.R. denotes Logical Reasoning, M.A. denotes math, S.T. denotes science and technology.}
\label{L_performance}
\setlength{\tabcolsep}{3pt}
\renewcommand{\arraystretch}{0.9}  % 调整行间距
\begin{tabular}{lccccccc}
\toprule
& \textbf{C.P.} & \textbf{F.P.} & \textbf{I.R.} & \textbf{L.R.} & \textbf{M.A.}  & \textbf{S.T.} & \textbf{Average}\\
\midrule
$\mathcal L=\{6\text{--}10\}$                      
    & 67.6 
    & 28.0
    & 38.7
    & 33.5
    & 29.6
    & 23.9
    & 36.9\\
$\mathcal L=\{5\text{--}11\}$       
    & 67.0
    & 27.6
    & 38.3
    & 33.2
    & 28.4
    & 23.7
    & 36.3\\
$\mathcal L=\{4\text{--}12\}$       
    & 67.0
    & 27.8
    & 39.5
    & 32.3
    & 29.0
    & 24.5
    & 36.7\\
$\mathcal L=\{7\text{--}9\}$        
    & 67.7
    & 28.0
    & 39.0
    & 32.2
    & 29.0
    & 23.3
    & 36.5\\
\bottomrule
\end{tabular}
\end{table}

\begin{table}[t]
\centering
\small
\caption{Ablation results of $\mathcal H$. C.P. denotes coarse perception, F.P. denotes fine-grained perception, I.R. denotes Instance Reasoning, L.R. denotes Logical Reasoning, M.A. denotes math, S.T. denotes science and technology.}
\label{H_performance}
\setlength{\tabcolsep}{3pt}
\renewcommand{\arraystretch}{0.8}  % 调整行间距
\begin{tabular}{lccccccc}
\toprule
& \textbf{C.P.} & \textbf{F.P.} & \textbf{I.R.} & \textbf{L.R.} & \textbf{M.A.}  & \textbf{S.T.} & \textbf{Average}\\
\midrule
$\mathcal H=\{13\text{--}21\}$                      
    & 65.8
    & 27.6
    & 38.3
    & 34.1
    & 29.6
    & 26.1
    & 36.9\\
$\mathcal H=\{22\text{--}24\}$       
    & 65.7
    & 28.5
    & 39.3
    & 34.1
    & 30.3
    & 25.0
    & 37.1\\
$\mathcal H=\{22\}$       
    & 66.1
    & 29.8
    & 39.4
    & 34.5
    & 30.3
    & 24.5
    & 37.4\\
$\mathcal H=\{23\}$        
    & 64.8
    & 29.3
    & 39.7
    & 34.2
    & 30.3
    & 24.3
    & 37.1\\
$\mathcal H=\{24\}$        
    & 65.1
    & 29.3
    & 39.4
    & 33.8
    & 30.0
    & 24.2
    & 37.0\\
\bottomrule
\end{tabular}
\end{table}

\subsubsection{Hyperparameter Ablation}
We determine the values of HERO’s hyperparameters ($\mathcal{L}$, $\mathcal{H}$, and $\alpha$) using LLaVA-NeXT-7B as the baseline HR-LVLM (with a 20\% token budget) and MMStar as the benchmark. MMStar is chosen for two reasons: (1) its well-defined evaluation dimensions enable task-dependent analysis of hyperparameter effects, and (2) it is independent of the benchmarks used in the main results, making it a fair validation set for hyperparameter selection.

\begin{figure}[t]
\centering
\includegraphics[width=0.4\textwidth]{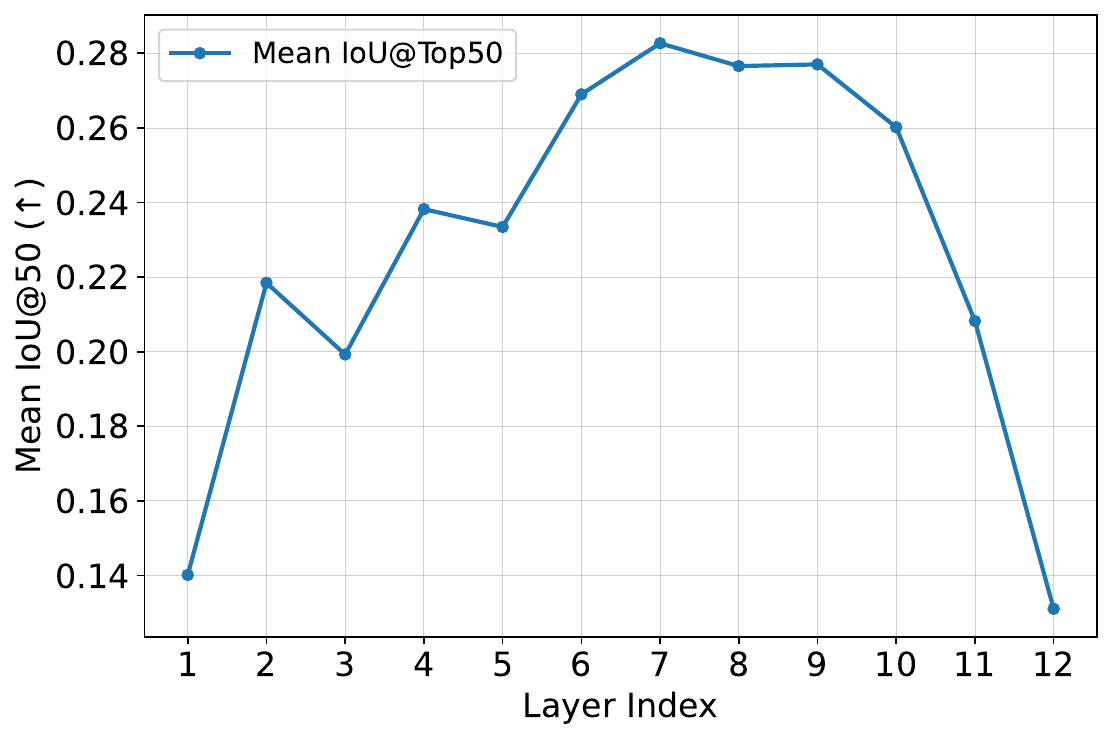}
\caption{Mean IoU between the top-50 CLS-attended patch tokens and salient-object masks across different layers of CLIP-ViT-L/336px, evaluated on the DUTS dataset.}
\label{IoU}
\end{figure}

We first explore the optimal setting for $\mathcal L$. 
As shown in Fig.~\ref{attention_similarity}(a), the CLS attention in the second stage (i.e., layers 13–24 in CLIP-ViT-L/336px) exhibits relatively uniform patterns. 
We therefore temporarily fix $\mathcal H=\{13\text{--}24\}$ and investigate which choice of $\mathcal L$ best captures primary tokens (i.e., tokens spatially aligned with the main semantic content of the image). 
To obtain an objective signal, we employ the DUTS salient-object detection dataset~\cite{duts}. 
For each layer in $\{1\text{--}12\}$, we select the top-50 patches by CLS attention and compute their mean IoU with the salient-object mask. 
As shown in Fig.~\ref{IoU}, layers 6-10 yield substantially higher IoU than other depths, indicating tighter alignment with foreground objects. 
This trend is consistent with the layerwise dynamics discussed in Section~\ref{subsection_attn_dynamics}. 
Guided by this observation, we set $\mathcal L=\{6\text{--}10\}$. 
We further validate this choice by expanding or shrinking the band and comparing their downstream performance. 
As shown in Table~\ref{L_performance}, $\mathcal L=\{6\text{--}10\}$ achieves the highest overall average, outperforming both wider bands such as $\{5\text{--}11\}$ and $\{4\text{--}12\}$, as well as the narrower band $\{7\text{--}9\}$. 
Therefore, we adopt it as the default setting for $\mathcal L$.

After fixing $\mathcal L=\{6\text{--}10\}$, we next explore the optimal setting for $\mathcal H$. As observed in Fig.~\ref{attention_similarity}(a), while CLS attention appears relatively stable across stage-2, the last three layers (22–24) show slight divergences. We therefore first compare two candidate ranges: $\mathcal H=\{13\text{--}21\}$ versus $\mathcal H=\{22\text{--}24\}$. As shown in Table~\ref{H_performance}, the latter achieves higher average performance, suggesting that the final three layers encode more informative cues for the thumbnail. To further disambiguate the roles of these layers, we evaluate them individually. Among $\mathcal H=\{22\}$, $\{23\}$, and $\{24\}$, the setting $\{22\}$ achieves the best overall score, outperforming both $\{23\}$ and $\{24\}$. This indicates that while all three final layers contribute global semantics, layer 22 provides the most effective balance between capturing high-level context and avoiding over-specialization observed in the very last layers. Based on these findings, we adopt $\mathcal H=\{22\}$ as the default setting for HERO.

\begin{figure}[t]
\centering
\includegraphics[width=0.4\textwidth]{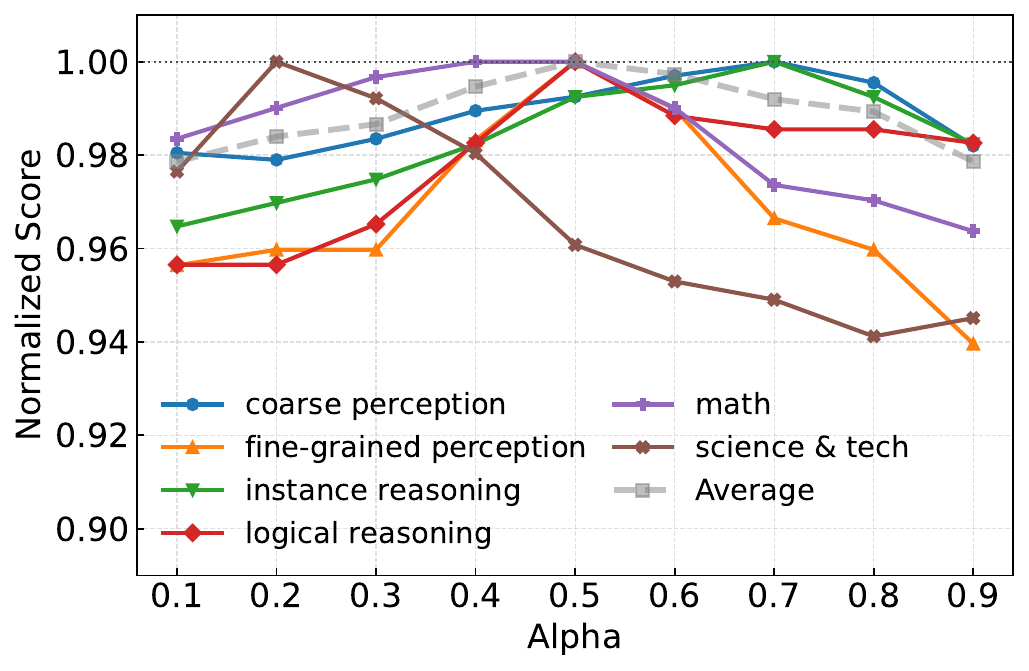}
\caption{Ablation results of the weighting coefficient $\alpha$ on MMStar.}
\label{alpha_ablation}
\end{figure}

We finally investigate the impact of the weighting coefficient $\alpha$, which balances visual saliency and textual relevance in token budget allocation.  
As shown in Fig.~\ref{alpha_ablation}, different task categories exhibit distinct preferences. 
For coarse perception and instance reasoning, larger values of $\alpha$ yield the best results, with both tasks peaking around $\alpha=0.7$, indicating that these tasks benefit more from emphasizing visually salient regions. 
In contrast, fine-grained perception, logical reasoning, and math follow an inverted-U trend, all reaching their peak at $\alpha=0.5$, while either increasing or decreasing $\alpha$ leads to performance drops, suggesting that these tasks require a balanced integration of visual and textual signals. 
Science \& technology shows the opposite tendency, achieving its highest score at $\alpha=0.2$ and steadily declining with larger $\alpha$, which highlights its stronger reliance on textual relevance. 
Overall, $\alpha=0.5$ offers the best trade-off across categories, achieving the highest average performance, and is therefore adopted as the default setting of HERO.

\begin{table*}[ht]
\centering
\small
\caption{Practical efficiency gains achieved by HERO under different token budgets on the MME benchmark.}
\label{computation}
\setlength{\tabcolsep}{5.5pt}
\renewcommand{\arraystretch}{0.9}  % 调整行间距
\begin{tabular}{lccccccc}
\toprule
\textbf{Method (Token Budget)} & \textbf{\# Tokens $\downarrow$} & \textbf{\parbox{2cm}{\centering TFLOPs $\downarrow$}} & \textbf{\parbox{2cm}{\centering GPU Memory \\(GB) $\downarrow$}} & \textbf{\parbox{2cm}{\centering KV-Cache \\(MB) $\downarrow$}} & \textbf{\parbox{2cm}{\centering Decoding Speed\\(tokens/sec) $\uparrow$}} & \textbf{Performance $\uparrow$} \\
\midrule
LLaVA-NeXT-7B(100\%)    & 2673 & 38.4 & 21.7 & 1336.5 & 28.8 & 1519.0 \\
HERO(40\%)      & 1105 & 14.1 & 17.4 & 552.5  & 35.4 & 1550.8 \\
HERO(20\%)      & 583  & 7.7  & 17.1  & 291.5  & 40.2 & 1500.9 \\
HERO(10\%)      & 322 & 4.2 & 16.9 & 161.0 & 43.8 & 1436.6 \\
\bottomrule
\end{tabular}
\end{table*}

\subsection {Practical Efficiency Analysis}
To assess the real-world efficiency gains brought by HERO, we evaluate its inference efficiency under different token budgets. We use LLaVA-NeXT-7B as the baseline LVLM and MME as the inference task. We conduct each evaluation with a batch size of 1 on a single GPU to eliminate the effects of parallelization and fairly reflect single-device inference efficiency. The results are summarized in Table \ref{computation}.

Compared with the full-token baseline, HERO substantially reduces computational cost while maintaining competitive performance. At the moderate 40\% budget, the number of input tokens decreases from 2673 to 1105, resulting in a 63\% reduction in TFLOPs (38.4 $\rightarrow$ 14.1), a 20\% reduction in GPU memory usage (21.7 GB $\rightarrow$ 17.4 GB), and a 59\% reduction in KV-cache size (1336.5 MB $\rightarrow$ 552.5 MB). Meanwhile, decoding speed improves from 28.8 to 35.4 tokens/sec, and performance surpasses the baseline (1550.8 vs.\ 1519.0). With a more aggressive 20\% budget, TFLOPs drop further to 7.7 and decoding speed increases to 40.2 tokens/sec, while performance remains competitive at 1500.9. 
At an extreme 10\% budget, HERO achieves 4.2 TFLOPs and 43.8 tokens/sec decoding speed, reducing cache consumption to less than one-seventh of the baseline, though at the cost of some performance degradation. These results demonstrate that HERO not only provides theoretical efficiency advantages but also translates them into end-to-end gains in practice. 

\subsection {Case Study}

To further illustrate the effectiveness of HERO, we present qualitative examples of its token pruning behavior under a visual token retention ratio of 20\%.

\begin{figure}[ht]
\centering
\includegraphics[width=0.47\textwidth]{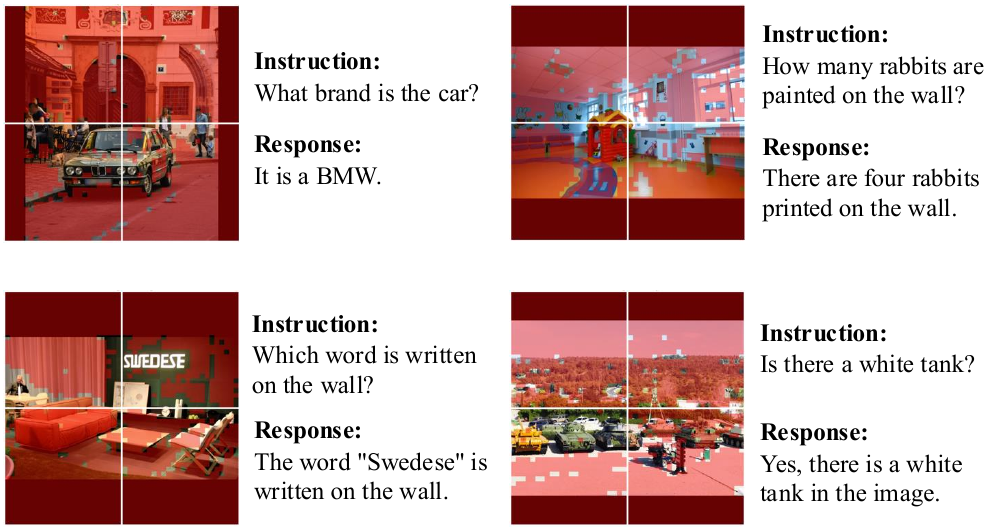}
\caption{Examples of HERO’s token pruning under a 20\% visual token budget. In each example, unmasked regions indicate the retained visual tokens.}
\label{case_study}
\end{figure}

As illustrated in Fig.\ref{case_study}, HERO consistently preserves both visually salient areas and task-relevant regions. For instance, when the instruction asks about the brand of a car, tokens covering the car are largely retained, while uninformative background regions are pruned. When the instruction concerns counting painted rabbits or reading specific text, HERO maintains tokens corresponding to these semantic objects or areas, even if they are spatially scattered or occupy a small portion of the image. Similarly, for questions about rare objects (e.g., a white tank), HERO successfully preserves the corresponding tokens despite the presence of surrounding visual clutter.

These visualizations demonstrate that HERO not only reduces the number of visual tokens, but does so in an instruction-aware and context-sensitive manner, effectively preserving the most critical information for visual understanding and downstream reasoning.

\section{Conclusion}

Motivated by insights into tile importance estimation, CLS attention dynamics, and specialized roles of visual tokens, we propose HERO, an inference-time visual token pruning strategy that effectively improves the computational efficiency of HR-LVLMs with minimal performance degradation. A potential limitation of HERO lies in its reliance on manually selected hyperparameters. While the default settings perform well on public benchmarks, the optimal values may vary across tasks. Future work will explore incorporating learnable hyperparameters to further enhance the adaptability and performance of HERO.

% \vfill

\bibliographystyle{unsrt}
\bibliography{aaai2026}

\appendix
\subsection{Evaluation Protocol for Image Captioning}
\label{appendix_caption}
\begin{figure}[ht]
\centering
\includegraphics[width=0.4\textwidth]{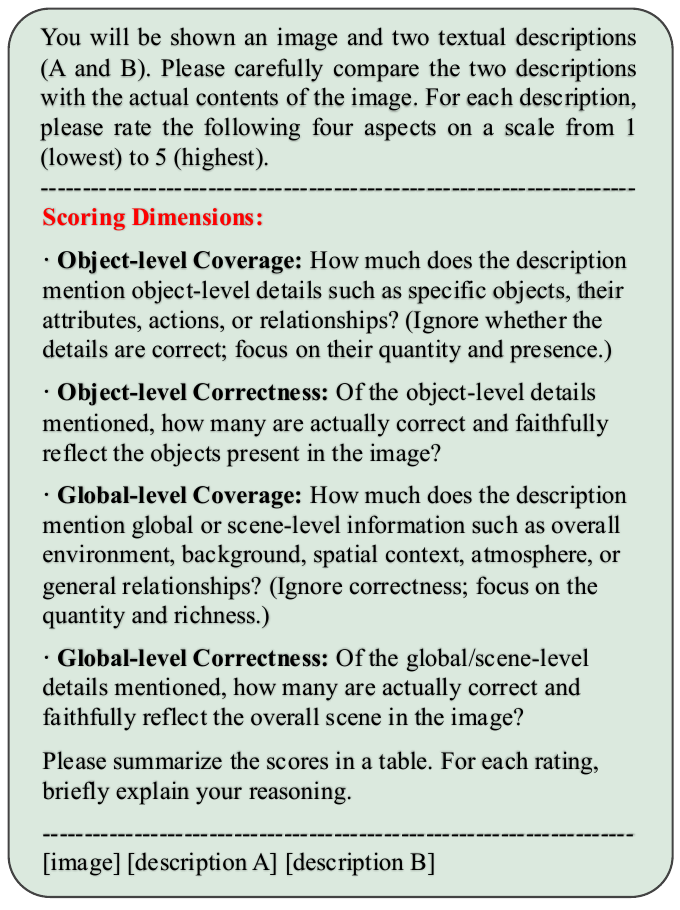}
\caption{The prompt provided to GPT-4o for image caption evaluation.}
\label{prompt}
\end{figure}

For each image in the MMStar benchmark, we employ LLaVA-1.5-7B to generate captions under two different inference settings: (1) retaining only the top 10\% CLS-attended patch tokens from layers 4–8, and (2) retaining only the top 10\% CLS-attended patch tokens from layers 12–23. This design allows us to contrast the descriptive power of primary tokens (emphasized in early-to-middle layers) with that of shortcut tokens (emphasized in middle-to-late layers). The generated captions are then evaluated using GPT-4o (API version: gpt-4o-2024-08-06), which compares them against the ground-truth image content and assigns quality scores along four predefined dimensions. The detailed scoring criteria and the evaluation prompt are provided in Fig.~\ref{prompt}.

\vfill
\subsection{Layerwise Visualizations of CLS Attention}
\label{appendix_layerwise}

\begin{figure}[ht]
\centering
\includegraphics[width=0.4\textwidth]{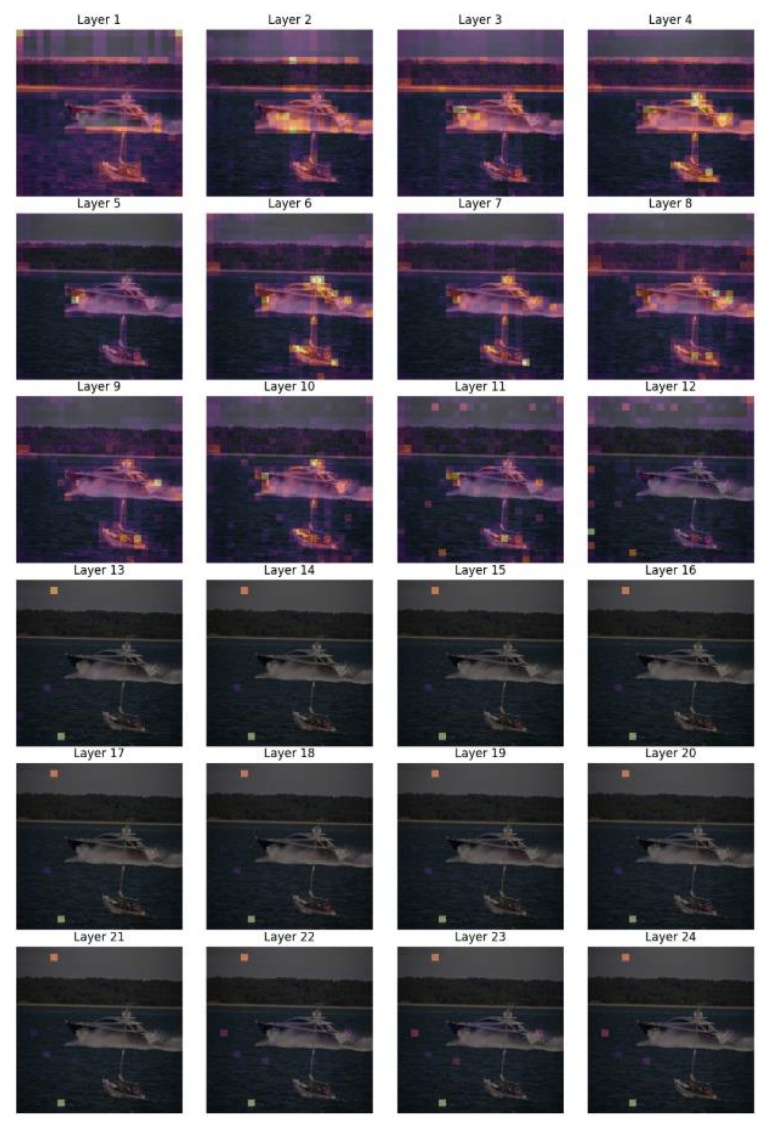}
\caption{Layerwise visualization of CLS attention on an image with clear foreground objects.}
\label{ship}
\end{figure}

\begin{figure}[h]
\centering
\includegraphics[width=0.4\textwidth]{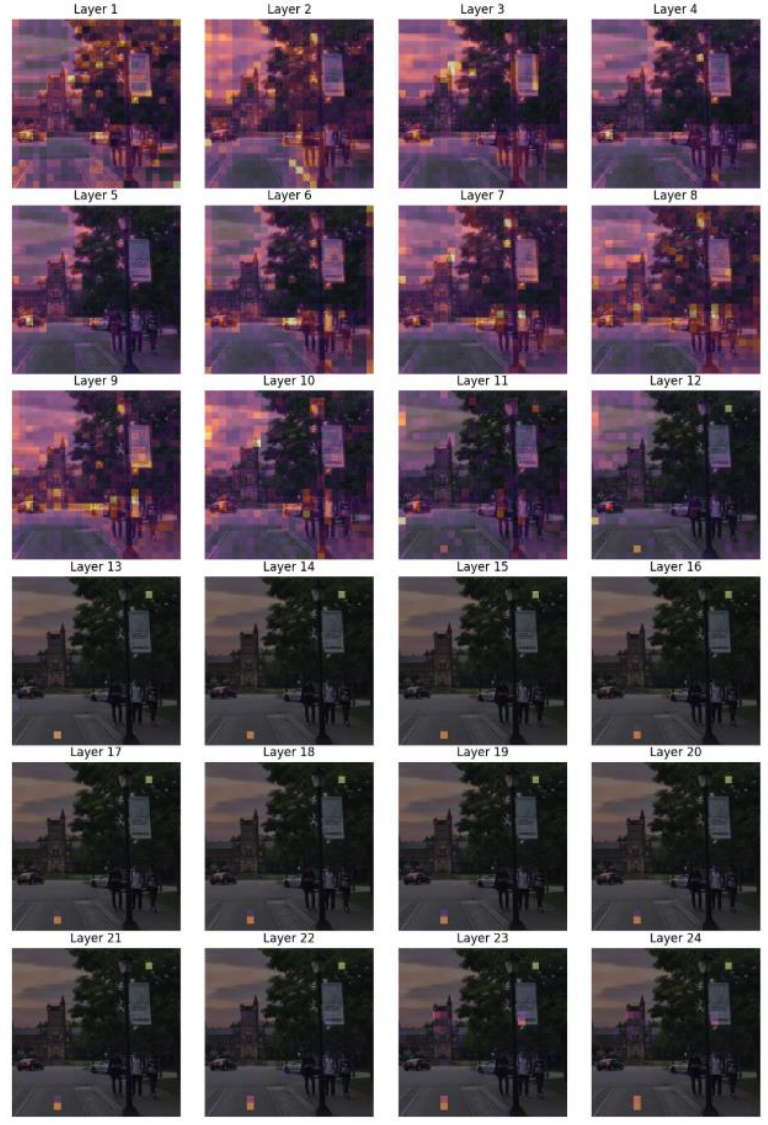}
\caption{Layerwise visualization of CLS attention on an image with less salient foreground objects and a cluttered background.}
\label{street}
\end{figure}

\end{document}